\documentclass[10pt,twocolumn,letterpaper]{article}

\usepackage{wacv}
\usepackage{times}
\usepackage{epsfig}
\usepackage{graphicx}
\usepackage{amsmath}
\usepackage{amssymb}
\usepackage{booktabs}
\makeatletter
\@namedef{ver@everyshi.sty}{}
\makeatother
\usepackage{tikz}

\usepackage{times}
\usepackage{epsfig}
\usepackage{graphicx}
\usepackage{amsmath}
\usepackage{amssymb}
\usepackage{comment}
\usepackage{bm}
\usepackage{xcolor}

\usepackage{comment}
\usepackage{multirow}
\usepackage[subrefformat=parens]{subcaption}
\usepackage{colortbl}

\def\wacvPaperID{384} 

\wacvalgorithmstrack   

\wacvfinalcopy 

\ifwacvfinal
\usepackage[breaklinks=true,bookmarks=false,colorlinks]{hyperref}
\else
\usepackage[pagebackref=true,breaklinks=true,colorlinks,bookmarks=false]{hyperref}
\fi

\usepackage[capitalize]{cleveref}
\crefname{section}{Sec.}{Secs.}
\Crefname{section}{Section}{Sections}
\Crefname{table}{Table}{Tables}
\crefname{table}{Tab.}{Tabs.}

\begin{document}

\title{nLMVS-Net: Deep Non-Lambertian Multi-View Stereo}

\author{Kohei Yamashita \qquad\qquad Yuto Enyo \qquad\qquad Shohei Nobuhara \qquad\qquad Ko Nishino\\
Graduate School of Informatics, Kyoto University,
Kyoto, Japan\\
{\tt\small \url{https://vision.ist.i.kyoto-u.ac.jp/}}
}

\twocolumn[{
\maketitle
\begin{center}
    \captionsetup{type=figure}
    \includegraphics[width=\linewidth]{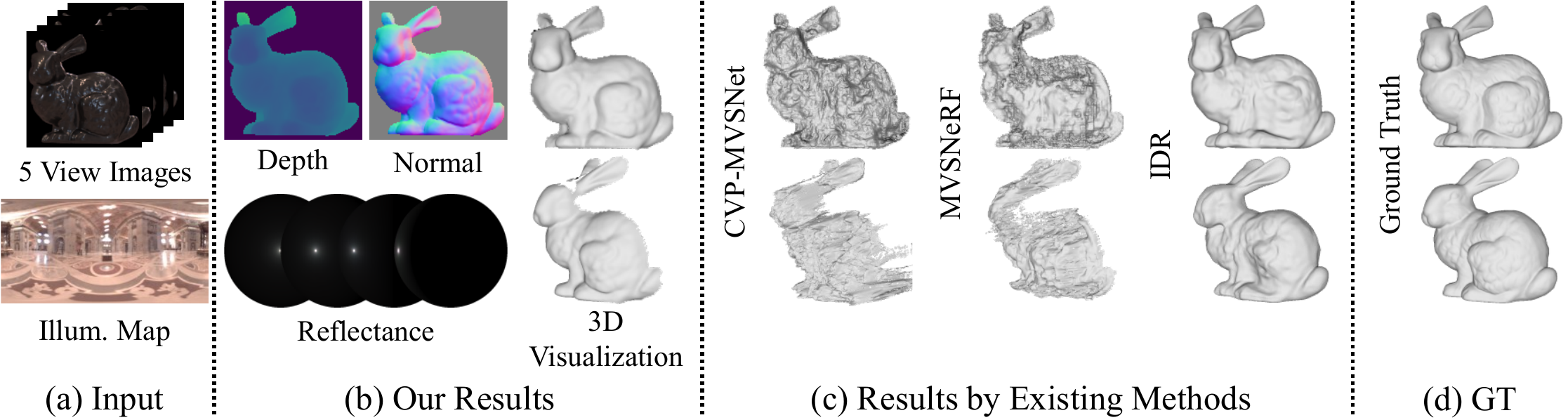}
        \captionof{figure}{We introduce a novel MVS method that can jointly estimate per-pixel depths and surface normals together with the complex reflectance (b) of a textureless object from five images around the view of interest captured under known but natural illumination (a). The example results show that our method successfully recovers 3D shape consistent with ground truth (d) from a sparse set of images from which existing MVS and neural view synthesis methods struggle to recover accurate geometry (c). Please see, for example, the face and legs of the Stanford Bunny.}
    \label{fig:opening}
\end{center}
}]
\thispagestyle{empty}

\begin{abstract}
\vspace{-12pt}
We introduce a novel multi-view stereo (MVS) method that can simultaneously recover not just per-pixel depth but also surface normals, together with the reflectance of textureless, complex non-Lambertian surfaces captured under known but natural illumination. Our key idea is to formulate MVS as an end-to-end learnable network, which we refer to as nLMVS-Net, that seamlessly integrates radiometric cues to leverage surface normals as view-independent surface features for learned cost volume construction and filtering. It first estimates surface normals as pixel-wise probability densities for each view with a novel shape-from-shading network. These per-pixel surface normal densities and the input multi-view images are then input to a novel cost volume filtering network that learns to recover per-pixel depth and surface normal. The reflectance is also explicitly estimated by alternating with geometry reconstruction. Extensive quantitative evaluations on newly established synthetic and real-world datasets show that nLMVS-Net can robustly and accurately recover the shape and reflectance of complex objects in natural settings.
\end{abstract}

\section{Introduction}

Three-dimensional reconstruction of real-world objects of arbitrary reflectance is essential for many computer vision applications. In particular, a passive approach that can recover the 3D geometry for a view from a handful of images would be preferable. For downstream tasks such as scene navigation, object grasping, and augmented reality, explicit recovery of the reflectance properties in addition to the 3D geometry would be essential. As shown in \cref{fig:opening}(c), these requirements are hard to fulfill with neural view synthesis methods (\eg, neural radiance field (NeRF)) as explicit geometry reconstruction is not their primary goal (\ie, volume density only provides coarse view-dependent surface geometry) and as they usually require dense view sampling. Classic stereopsis and multi-view stereo (MVS) approaches \cite{scharstein2002taxonomy}, especially with recent integration of learned features and filtering, still excel in their simplicity, accuracy, and passive setup for explicit 3D geometry reconstruction.

Reconstruction of a textureless non-Lambertian surface (\eg, a porcelain vase), however, still remains elusive to stereo-based approaches as stereopsis is limited by its two fundamental requirements: correspondence matching and triangulation. Finding correspondences directly translates to making assumptions about the surface appearance, that they can be matched across views, \ie, they are view-independent and texture-rich. This has largely limited the application of stereo-based methods to textured and Lambertian surfaces. Geometry recovery by triangulation also limits the output to surface depth which is often insufficient for capturing details of surface geometry. 

Surface geometry can instead be recovered as part of inverting the radiometric process of image formation. Various methods have been proposed for single-view geometry reconstruction of non-Lambertian surfaces by jointly estimating its reflectance. The geometry recovered in such inverse rendering approaches is, however, fundamentally limited to surface normals. Although surface normals can expose finer surface details, they do not directly represent the surface. 

In this paper, we introduce a novel multi-view stereo method that enables the simultaneous recovery of surface normals and depth for textureless non-Lambertian surfaces. At the same time, the method explicitly recovers the complex reflectance of the target surface. Our method is purely image-based. As we experimentally show in \cref{sec:results}, it requires only a handful of (\ie, 5) neighboring views for reconstruction from one vantage point. Most important, our method can be applied to objects with unknown complex reflectance captured under known but natural illumination. Our key idea is to integrate stereopsis with radiometric analysis so that radiometrically recovered geometric properties, namely surface normals, can serve as view-independent cues for multi-view stereopsis. We achieve this integration with an end-to-end learnable network which we refer to as nLMVS-Net.

Our nLMVS-Net consists of three key novel ideas. The first is a single-view shape-from-shading network that fully leverages radiometric likelihoods of surface normals. The network enables the estimation of per-pixel surface normal as a directional probability density which collectively serves as rich view-independent cues for subsequent multi-view stereo. The second key idea is a novel cost volume filtering network that leverages the recovered surface normal probability densities. The network integrates radiometric (\ie, surface normals) and geometric (\ie, correspondences) cues with a novel feature extraction layer and a consistency loss between the surface normal and depth estimates. The third is joint estimation of complex non-Lambertian reflectance by alternating with geometry estimation using a neural BRDF model \cite{chen2020ibrdf}.

\begin{table}[t]
  \centering
  \setlength{\tabcolsep}{2.7pt}
  \scriptsize
  \begin{tabular}{l|cccccc}
     & Depth & Normal & 
     \begin{tabular}{c} 
        Reflectance\\Estimation
     \end{tabular} & 
     \begin{tabular}{c} 
        Sparse \\ View \\ Input
     \end{tabular} & 
     \begin{tabular}{c} 
        Nat. \\ Illum.
     \end{tabular} & 
     \begin{tabular}{c} 
        General \\ Object \\ Category
     \end{tabular} \\ \hline
    MVSNet \cite{yao2018mvsnet} & \checkmark &  & & \checkmark & \checkmark & \checkmark \\
    CVP-MVS \cite{yang2020cvpmvsnet} & \checkmark & & & \checkmark & \checkmark & \checkmark \\
    NAS \cite{kusupati2020nas} & \checkmark & \checkmark & & \checkmark & \checkmark & \checkmark \\ 
    RC-MVS \cite{chang2022rcmvsnet} & \checkmark & & & \checkmark & \checkmark & \checkmark \\
    \hline
    Nam \etal \cite{nam2018svbrdf} & \checkmark & \checkmark & \checkmark & &  & \checkmark \\
    Kaya \etal \cite{kaya2022umvps} & \checkmark & \checkmark & & & & \checkmark \\
    Cheng \etal \cite{cheng2021mv3d} & \checkmark & \checkmark & \checkmark & \checkmark & & \checkmark \\
    Bi \etal \cite{bi2020deep3d} & \checkmark & \checkmark & \checkmark & \checkmark &  & \checkmark \\
    ON \cite{oxholm2015shape} & \checkmark & \checkmark & \checkmark & & \checkmark & \checkmark \\
    \hline
    NeRFactor \cite{zhang21nerfactor} & \checkmark & \checkmark & \checkmark & & \checkmark & \checkmark \\
    PhySG \cite{zhanf2021physg} & \checkmark & \checkmark & \checkmark & & \checkmark & \checkmark \\
    IDR \cite{yariv2020idr} & \checkmark & \checkmark & & & \checkmark & \checkmark \\
    NeuS \cite{wang2021neus} & \checkmark & \checkmark & & & \checkmark & \checkmark \\
    NeRS \cite{zhang2021ners} & \checkmark & \checkmark & \checkmark & \checkmark & \checkmark & \\
    MVSNeRF \cite{chen2021mvsnerf} & \checkmark & & & \checkmark & \checkmark & \checkmark \\
    \hline
    \rowcolor[rgb]{0.93,1.0,0.87} Ours & \checkmark & \checkmark & \checkmark & \checkmark & \checkmark & \checkmark
  \end{tabular}
  \caption{Image-based 3D reconstruction methods that exploit multi-view observations. Our method can recover geometry and reflectance from sparse (5 views) observations captured under known but natural illumination without any category-specific shape prior, which remains challenging to existing methods.}
  \label{tab:methods}
\end{table}

\begin{figure*}[t]
  \centering
  \includegraphics[width=\linewidth]{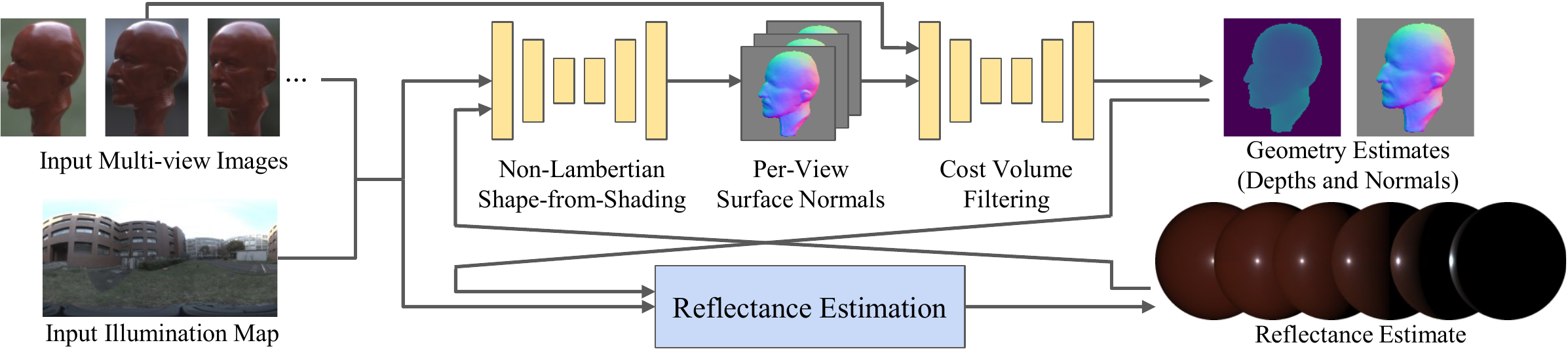}
  \caption{Overview of our novel multi-view stereo method, nLMVS-Net. The shape-from-shading sub-network learns to recover per-pixel probability densities of surface normals for each view. The novel cost volume sub-network then learns to reconstruct per-pixel depth and surface normals from those and the input sparse multi-view images. By alternating with neural reflectance estimation, we jointly recover the complex surface reflectance of the object.
  }
  \label{fig:overview}
\end{figure*}

We also introduce two newly collected datasets, which we refer to as nLMVS-Synth and nLMVS-Real. The synthetic dataset (nLMVS-Synth) consists of 26850 rendered images of 2685 objects with 94 and 2685 different real-world reflectance and natural illumination, respectively. We use nLMVS-Synth to train nLMVS-Net and thoroughly evaluate its accuracy on unseen synthetic images. The new multi-view image dataset of real objects, namely nLMVS-Real, consists of 2569 multi-view images of 5 objects each with one of 4 different reflectances taken under 6 different natural illuminations. Each of the 5 different objects is replicated with a 3D printer so that accurate ground truth geometry is available for quantitative analysis. This dataset is unprecedented in size for an accurately radiometrically and geometrically calibrated multi-view image set for a variety of surfaces and would undoubtedly serve as a useful platform for a wide range of shape reconstruction and inverse rendering research.

Experimental results on these datasets and others, together with direct comparisons with existing methods, clearly demonstrate the effectiveness of our method. We also show that the recovered depths and surface normals can be used to reconstruct a whole object from as few as 10 images. Thanks to the passive setup and sparse inputs, nLMVS-Net may prove useful for many 3D sensing applications including mobile sensing, XR immersion, and robotic navigation. All the data and code are publicly disseminated on our project web page.

\section{Related Work}
\label{sec:related_work}
We review relevant works on imaged-based 3D geometry reconstruction, mainly those methods that exploit multi-view observations. \Cref{tab:methods} summarizes the differences of our method and others. Our method can recover geometry and reflectance of textureless, non-Lambertian surfaces from a sparse set (\ie, 5) of multi-view images captured under known but natural illumination without any category-specific shape prior, which remains challenging for existing methods.

\textbf{Multi-view stereo} relies on cross-view correspondence matching and triangulation. Traditional methods relied on manually designed distance metrics for correspondence detection and spatial aggregation \cite{rhemann2011fastcostvolume, galliani2015gipuma}. Recent works leverage deep neural networks to learn the metric directly from data. In particular, 3D convolutional neural networks are often used for cost volume filtering \cite{yao2018mvsnet, im2019dpsnet, yao2019rmvsnet, yang2020cvpmvsnet}. Kusupati \etal \cite{kusupati2020nas} trained a deep neural network with a consistency loss to jointly estimate per-pixel depth and surface normal. The architecture of the cost volume filtering network of nLMVS-Net is inspired by this work, but fundamentally differs in that it handles not only textureless surfaces, but also non-Lambertian reflectance and even explicitly estimates it.  Chang \etal \cite{chang2022rcmvsnet} introduced a view synthesis loss that can implicitly handle non-Lambertian appearance. Their method, however, still relies on a photometric consistency loss which cannot handle large deviations from Lambertian reflectance.

\textbf{Inverse rendering} methods invert radiometric image formation to reconstruct object geometry \cite{horn1970sfs, johnson11sfs, barron2015sirfs, hertzmann2005examplebased, nam2018svbrdf}. Multi-view methods have exploited proxy geometry (\eg, a 3D mesh model) to recover surface geometry by iterative, nonlinear optimization \cite{oxholm2015shape, xia2016xfm, nam2018svbrdf, zhanf2021physg, kaya2022umvps, kaya2022neuralmvps}. Oxholm and Nishino \cite{oxholm2015shape} alternated between updating a 3D mesh and BRDF parameters so that they are consistent with the input multi-view images and a known illumination map. These approaches struggle to recover high-frequency details of surface geometry as nonlinear optimization of geometry is unstable due to the large number of free parameters, especially when the number of input images is small (\eg, 20). A few methods handle sparse-view inputs \cite{bi2020deep3d, cheng2021mv3d}. They, however, require collocated point lighting. In contrast, our method works with sparse inputs (5 images and an illumination map) under complex natural illumination. 

\textbf{Neural image synthesis} methods recover a volumetric representation of a scene from a large number (typically on the order of tens to 100) of multi-view images \cite{zhang21nerfactor, zhanf2021physg, boss2021neuralpil, yariv2020idr, wang2021neus}. Chen \etal \cite{chen2021mvsnerf} handled sparse (\ie, 3) views by conditioning the volume representation with input images. This method relies on volume rendering and the recovered volume density only provides view-dependent coarse depths.  Zhang \etal \cite{zhang2021ners} achieved explicit reconstruction of surface geometry and reflectance from sparse (\ie, 7) views by leveraging category-specific shape templates. They also recovered 3D shapes for arbitrary object categories by exploiting cuboids as templates. As we show in the supplementary material, this method struggles to generalize to objects with complex geometry that cannot be approximated with cuboids. 

\textbf{Multi-view stereo datasets} have been proposed for benchmarking \cite{aanaes2016dtumvs, schoeps2017eth3d, Knapitsch2017tanksandtemples, yao2020blendedmvs}. They, however, are not radiometrically calibrated as they focus on Lambertian or textured surfaces rather than non-Lambertian objects. For the evaluation of non-Lambertian MVS, linear high dynamic range images that accurately capture the appearance of non-Lambertian objects are essential. Although the Multiview Objects under Natural Illumination database \cite{oxholm2014mvshape} provides high dynamic range images along with ground truth geometry and illumination maps, the number of instances is limited (4 objects under 3 environments). We also found that the images of this dataset contain flaws (please see \cref{sec:results}). We introduce a novel real-world dataset that is accurate and extensive which can serve as a new platform for further studies on shape and reflectance recovery.

\section{Deep Non-Lambertian Multi-View Stereo}

\Cref{fig:overview} depicts the overall structure of our model nLMVS-Net. The inputs are five multi-view images of an object and an illumination map of the surrounding environment. We assume the latter can be captured with a light probe, or can be estimated with a separate method \cite{LeGendre2019deeplight, gardner2017indoor, Zhang2019outdoorlight}. Our nLMVS-Net consists of a single-view shape-from-shading network and a cost volume filtering network. Since the appearance of non-Lambertian objects (\eg, gloss) changes according to the viewing direction, we cannot directly achieve correspondence matching on the input images, especially for textureless surfaces. Instead, we explicitly extract view-independent features, namely surface normals, but while canonically accounting for uncertainty by encoding them as directional distributions with the shape-from-shading network. The recovered per-pixel surface normal distributions add rich information in addition to the regular appearance for multi-view correspondence matching and shape reconstruction. We derive a novel cost-volume filtering network that achieves seamless integration of these rich geometric and visual cues. We also derive a joint estimation framework that explicitly estimates the surface reflectance expressed by an invertible network together with the normals and depth.

\subsection{Non-Lambertian Shape-from-Shading}
\label{sec:sfs}
\begin{figure}[t]
  \centering
  \subfloat[][Shape-from-Shading (SfS) Network]{
    \includegraphics[keepaspectratio, width=0.97\linewidth]{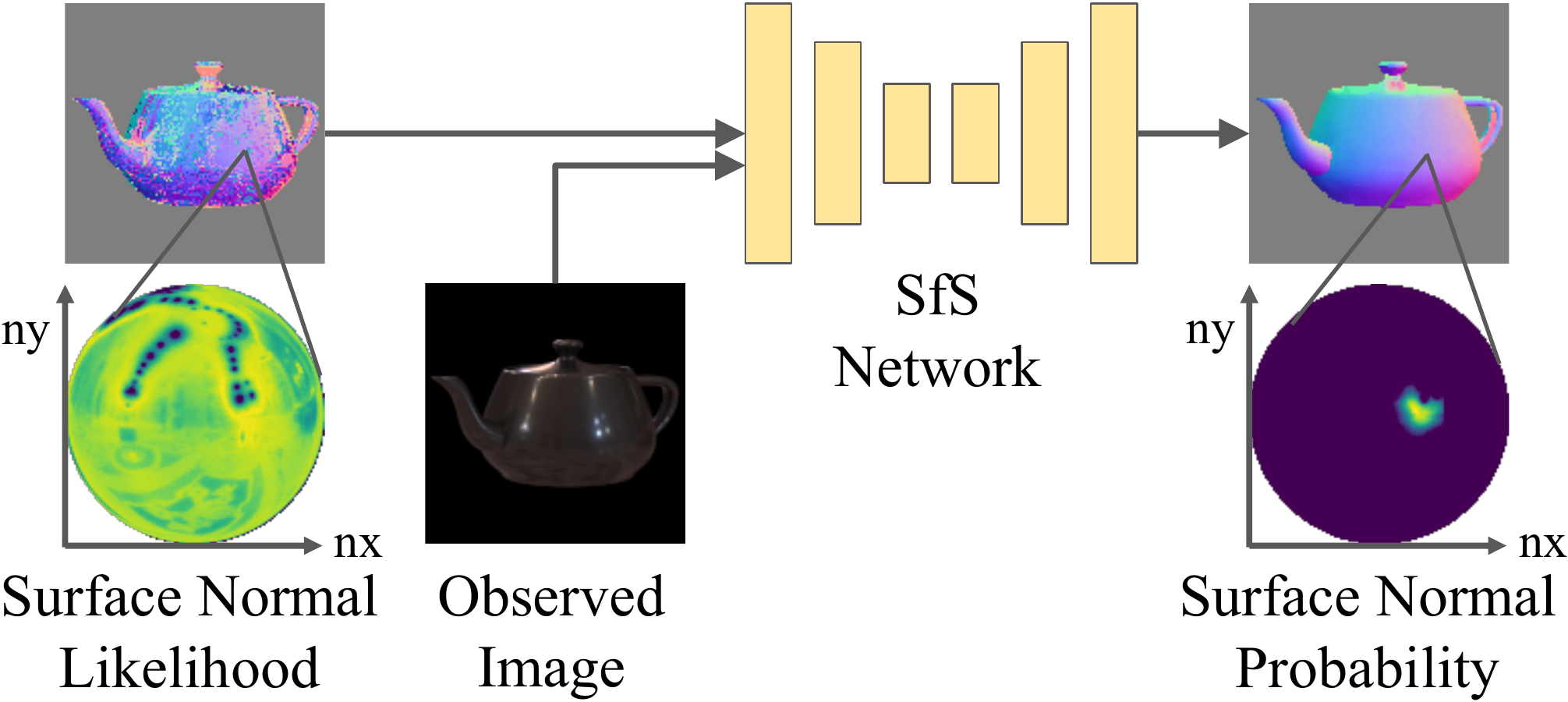}
    \label{fig:sfsnet_concept}
  }
  \\
  \vspace{6pt}
  \subfloat[][Coarse-to-Fine SfS]{
    \includegraphics[keepaspectratio, width=0.97\linewidth]{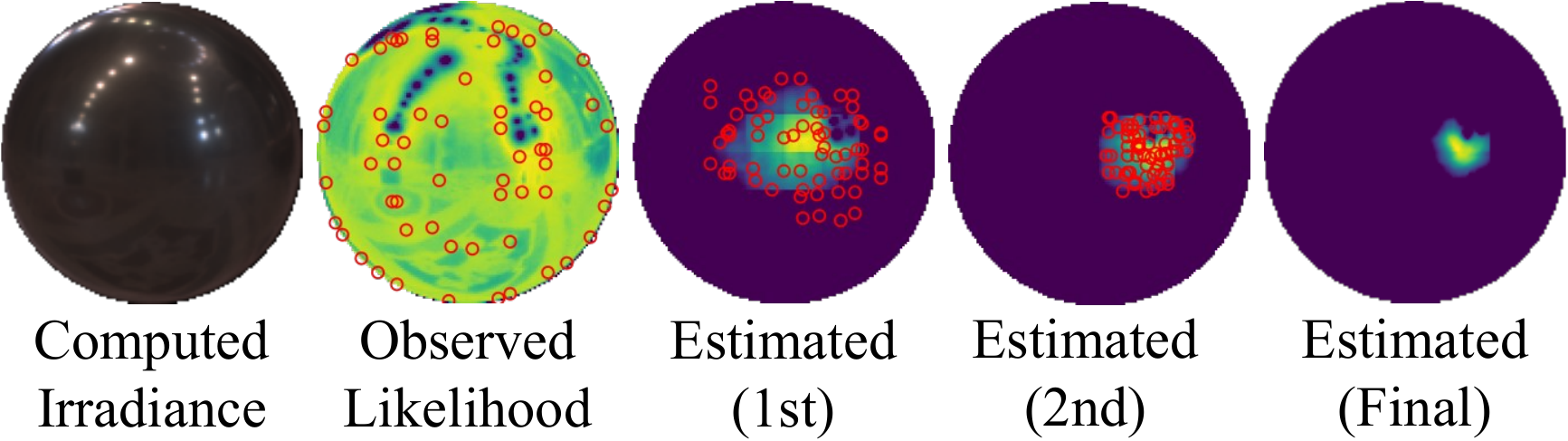}
    \label{fig:coarse-to-fine}
  }
   \caption{\subref{fig:sfsnet_concept} The shape-from-shading (SfS) network of nLMVS-Net learns to estimate pixel-wise probability densities of surface normals by aggregating local and global contextual information from the input view and observed pixel-wise likelihoods computed from the radiometric image formation model. \subref{fig:coarse-to-fine} We use the SfS network recursively to refine the observed likelihoods in a coarse-to-fine manner. The red circles on the probability densities are the sampled surface normal orientations, which are used as inputs to the network in subsequent iterations.}
   \label{fig:sfsnet}
\end{figure}

The first step of nLMVS-Net is to recover the surface normals with associated uncertainties for each view. Surface normals naturally lie in a plausible range of directions for a given intensity as neither the illumination nor the reflectance is angularly unique \cite{oxholm2014mvshape}. As such, it is essential to model their uncertainties. For this, as depicted in \cref{fig:sfsnet_concept}, we derive a novel deep neural network that estimates the per-pixel probability density distribution of surface normals for each view of the multi-view input images.

We assume an opaque, homogeneous reflectance for the object whose BRDF can be expressed as $\rho(\bm{\omega_i}, \bm{\omega_o}, \bm{n})$, where $\bm{\omega_i}$ is the incident direction, $\bm{\omega_o}$ is the viewing direction, and $\bm{n}$ is the surface normal. We also assume that the cameras and the illumination environment are distant from the object, \ie, they can be approximated with an orthographic camera and an illumination map $L_i(\bm{\omega_o})$. Under these assumptions, the observed irradiance $E(\bm{\omega_o}, \bm{n})$ is
\begin{equation}
    E(\bm{\omega_o}, \bm{n}) = \int L_i(\bm{\omega_i})\rho(\bm{\omega_i}, \bm{\omega_o}, \bm{n})\max(0, \bm{\omega_i}\cdot\bm{n})\mathrm{d}\bm{\omega_i}\,.
    \label{eq:image-formation}
\end{equation}
We leave global light transport including shadows and interreflections for future work, and focus on object appearance by direct lighting which is dominant for single objects.

\begin{figure*}[t]
  \centering
  \includegraphics[width=\linewidth]{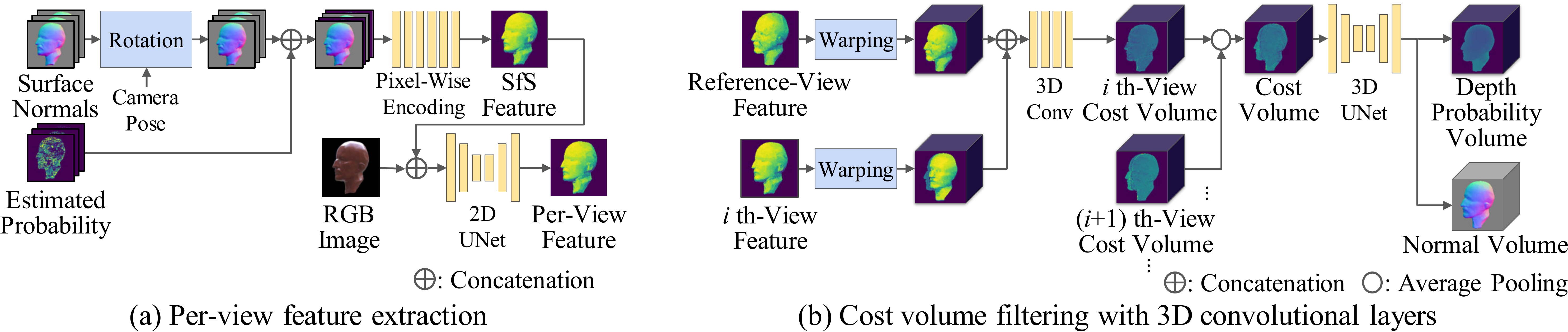}
  \caption{The architecture of the cost volume filtering network of nLMVS-Net. A latent cost volume is constructed from the multi-view surface normal densities and color appearance which is then filtered with 3D convolutional layers. The network outputs two 3D volumes: depth probability volume and surface normal volume that encode the per-pixel depth and surface normal estimates as probability densities.}
  \label{fig:mvsnet}
\end{figure*}

Let us assume that we are given a current estimate of the reflectance. This reflectance will be updated later in an alternating outer loop. For a given hypothesized surface normal and known illumination, its likelihood can be defined as the similarity of the irradiance $E(\bm{\omega_o},\bm{n})$ computed from the given normal and the actual pixel value $I$:
\begin{equation}
    p(I|\bm{n}) = \prod_k f(\log I^{(k)}; \log E^{(k)}(\bm{\omega_o},\bm{n}), b)\,,
    \label{eq:normallikelihood}
\end{equation}
where $f(x; \mu, b)$ is the Laplace distribution and $k$ is index of color channels. We use the Laplace distribution as its long tail is suitable for modeling deviations from the image formation model caused, for instance, by shadows and interreflections. We optimize the parameter $b$ with training data. The surface normal directions are discretized with a 2D hemispherical grid and the likelihoods are computed for each direction. 

The observed surface normal likelihoods \cref{eq:normallikelihood} are too noisy and unreliable to use for cost volume filtering. We train the shape-from-shading network to refine and convert them into probability density distributions by aggregating local and global contextual information across the surface. As depicted in \cref{fig:coarse-to-fine}, for computational efficiency, we achieve this in a coarse-to-fine manner. We first divide the possible surface normal orientations into a 8$\times$8 grid and, for each grid, find the orientation that maximizes the observed likelihood $p\left(I| \bm{n} \right)$ with brute-force search. We use the set of sampled surface normals and their observed likelihoods as inputs for each pixel. In the subsequent iterations, we double the resolution of the grid and sample surface normals around those that have high probability in the previous iteration. We use the same network with the same weights for all stages.

Since the inputs of the network (\ie, sets of surface normals and their observed likelihoods) are unstructured, convolutions are not suitable for processing them. Instead, inspired by PointNet \cite{qi2016pointnet}, we extract a 64 dimensional feature vector for each sampled surface normal and then fuse the sample-wise features by using max-pooling. The fused features are concatenated with features extracted from the input image and filtered by 2D convolutional layers. We use the pixel-wise filtered feature $\bm{a}$, a decoder MLP $g(\bm{n}; \bm{a})$ that outputs a scalar value, and the observed likelihood distribution $p(I|\bm{n})$ to compute the output (unnormalized) probability density distribution $\hat{p}(\bm{n})$ 
\begin{equation}
    \hat{p}(\bm{n}) = p(I|\bm{n}) g(\bm{n}; \bm{a})\,.
\end{equation}

We train the network with images of synthetic objects whose BRDF and surface normals are known. In training, we compute the observed surface normal likelihoods by using the ground-truth BRDF and evaluate the network output with cross entropy loss
\begin{equation}
    L_\textrm{SfS} = -\sum_{i}M(\bm{n_i}) \log \left(\frac{\hat{p_i}(\bm{n_i})}{\sum_j \hat{p_j}(\bm{n_j})}\right),
\end{equation}
where $\{\bm{n_i}\}$ is the sampled (input) surface normal direction and $M$ is a binary mask (1 iff the ground-truth and the sampled surface normal are in the same grid). The loss is evaluated for every stage of the coarse-to-fine estimation.

As shown in \Cref{fig:sfsnet_concept}, the network significantly reduces the ambiguity of the observed likelihood distribution and extracts a well-defined probability density for each pixel.

\subsection{Cost Volume Filtering}
\label{sec:cost_volume_filtering}

From the recovered per-pixel surface normal probability densities for each view as well as the original input images, nLMVS-Net learns to filter a cost volume to recover the object 3D shape as depth and surface normals. \Cref{fig:mvsnet} shows the architecture of the cost-volume filtering network.   

As depicted in \cref{fig:mvsnet}(a), the network takes in the multi-view input images as well as the outputs of the single-view shape-from-shading network, \ie, per-pixel surface normal probability densities for each view. The latter are represented as sets of surface normals and their probabilities, from which we extract pixel-wise features. Since the surface normals are estimated in the camera coordinate system of each view, we first consolidate the coordinate system by rotating them using the known camera extrinsic parameters. We apply a feature extraction layer similar to the one in the shape-from-shading network (see \cref{sec:sfs}) to convert the unstructured set of surface normals and their probabilities into a latent feature vector. The latent vector is concatenated with the input image and further filtered by a 2D UNet.

As illustrated in \cref{fig:mvsnet}(b), the surface normal and image features are then used to construct a 3D latent cost volume which is then filtered with 3D convolutional layers. The final outputs are per-pixel depths and surface normals in the reference view. The outputs of the 3D convolutional layers become a depth probability volume $\hat{p}(d;\bm{m})$ and a surface normal volume $\hat{\bm{n}}(d,\bm{m})$ where $d$ is a discretized hypothesis of depth. From these volumes, we compute the estimated depth $\hat{d}(\bm{m})$ and surface normal $\hat{\bm{n}}(\bm{m})$ as
\begin{equation}
    \hat{d}(\bm{m}) = \sum_d \hat{p}(d;\bm{m}) d\,,
\end{equation}
\begin{equation}
    \hat{\bm{n}}(\bm{m}) = \frac{\sum_d \hat{p}(d;\bm{m})\hat{\bm{n}}(d,\bm{m})}{\|\sum_d \hat{p}(d;\bm{m})\hat{\bm{n}}(d,\bm{m})\|}\,.
\end{equation}

\begin{figure}[t]
  \centering
  \includegraphics[keepaspectratio, width=\linewidth]{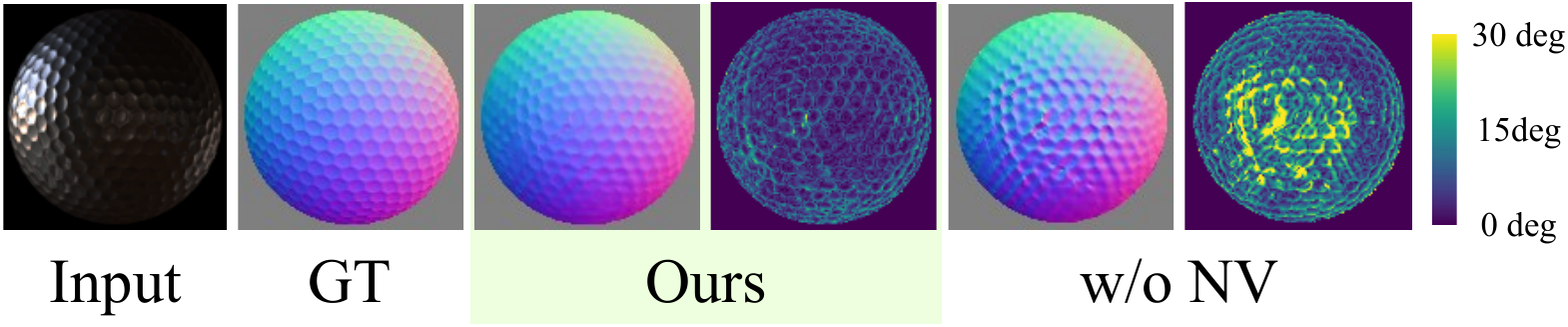}
  \caption{Advantage of estimating surface normals and depth as separate quantities. Our surface normal estimates are mostly consistent with the true surface normals (GT) in both continuous regions and also at surface discontinuities. In contrast, if depth derivatives (differentiation) are used as surface normals (\ie, nLMVS-Net without normal volume--``w/o NV''), surface normals become erroneous at even continuous regions and high-frequency details of surface discontinuities are also lost or exaggerated. }
  \label{fig:ablation_normal_volume}
\end{figure}

We ensure that the estimated depth and surface normals mostly agree with each other with a loss that aggregates the discrepancy in the directions of the surface normals and depth derivatives 
\begin{equation}
    L_\textrm{dn} = \sum_{\bm{m}}\arccos\left(\hat{\bm{n}}(\bm{m})\cdot\bar{\bm{n}}(\bm{m})\right),
    \label{eq:loss_consistency}
\end{equation}
where $\bar{\bm{n}}(\bm{m})$ is the surface normal computed from the estimated depths $\hat{d}(\bm{m})$ with cross product of tangent vectors on the surface. As shown in \cref{fig:ablation_normal_volume}, this consistency holds only in C0 and C1 continuous regions and we do not use $\bar{\bm{n}}(\bm{m})$ as surface normal estimates. The summation in \cref{eq:loss_consistency} ensures that strict consistency is only mildly imposed and surface normals are allowed to deviate from the depth derivatives at surface discontinuities. This is possible as we recover depths and surface normals as separate quantities.

We also impose individual depth and surface normal supervisions
\begin{equation}
    L_\textrm{d} = \sum_{\bm{m}} \|\hat{d}(\bm{m}) - d_\textrm{gt}(\bm{m})\|_1\,,
\end{equation}
\begin{equation}
    L_\textrm{n} = \sum_{\bm{m}} \arccos\left(\hat{\bm{n}}\left(\bm{m}\right) \cdot \bm{n_\textrm{gt}}(\bm{m}) \right)\,,
\end{equation}
where $d_\mathrm{gt}(\bm{m})$ and $\bm{n_\mathrm{gt}}(\bm{m})$ are the ground-truth depth and surface normal. The overall training loss is the weighted sum of these loss functions. We train this network separately from the shape-from-shading network.

\subsection{Joint Shape and Reflectance Estimation}
\label{sec:shape_and_reflectance}
As \cref{fig:overview} depicts, we alternate between estimating the object geometry and estimating the reflectance (BRDF). We represent the surface BRDF with the conditional invertible neural BRDF model (conditional iBRDF model) \cite{chen2020ibrdf}. We update parameters of this model so that the difference between the input view images and the rendered images for each view is minimized. A challenge here is that pixel-wise intensity errors are too brittle as they suffer from geometry reconstruction errors. For this, we derive two objective functions that explicitly handle the reconstruction errors. The key ideas are that 1) we blur the images to evaluate the consistency in a coarse level, and that 2) we can find the ground truth surface normal around the estimated one that almost exactly satisfies the radiometric consistency and use it for rendering. Please see the supplementary material for further details.

\subsection{Whole 3D shape Recovery}
\label{sec:whole_shape_recovery}
Our nLMVS-Net can recover per-pixel surface normal and depth of a reference view image from 5 input view images. If we have a set of such multi-view images that collectively cover the entire object (typically 10 images) taken while moving around an object on a plane, we can recover the whole 3D object geometry by applying nLMVS-Net to 5 images each while each image becomes the reference view, after which we integrate all depth and surface normals. During the alternating estimation of geometry and reflectance, we can use all the images together to update a single reflectance estimate. In the geometry estimation, for each view, we select four neighboring views (2 to the left and 2 to the right for a typical 360$^\circ$ capture on a plane) as inputs to nLMVS-Net. In the reflectance estimation, we compute the objective function introduced in \cref{sec:shape_and_reflectance} for each view and minimize the sum of them. We then reconstruct a 3D mesh from the estimated depth and surface normals by converting them into oriented points and applying Poisson surface reconstruction \cite{kazhdan2006poisson}.

\section{Experimental Results}
\label{sec:results}

\begin{figure}[t]
  \centering
  \includegraphics[keepaspectratio, width=\linewidth]{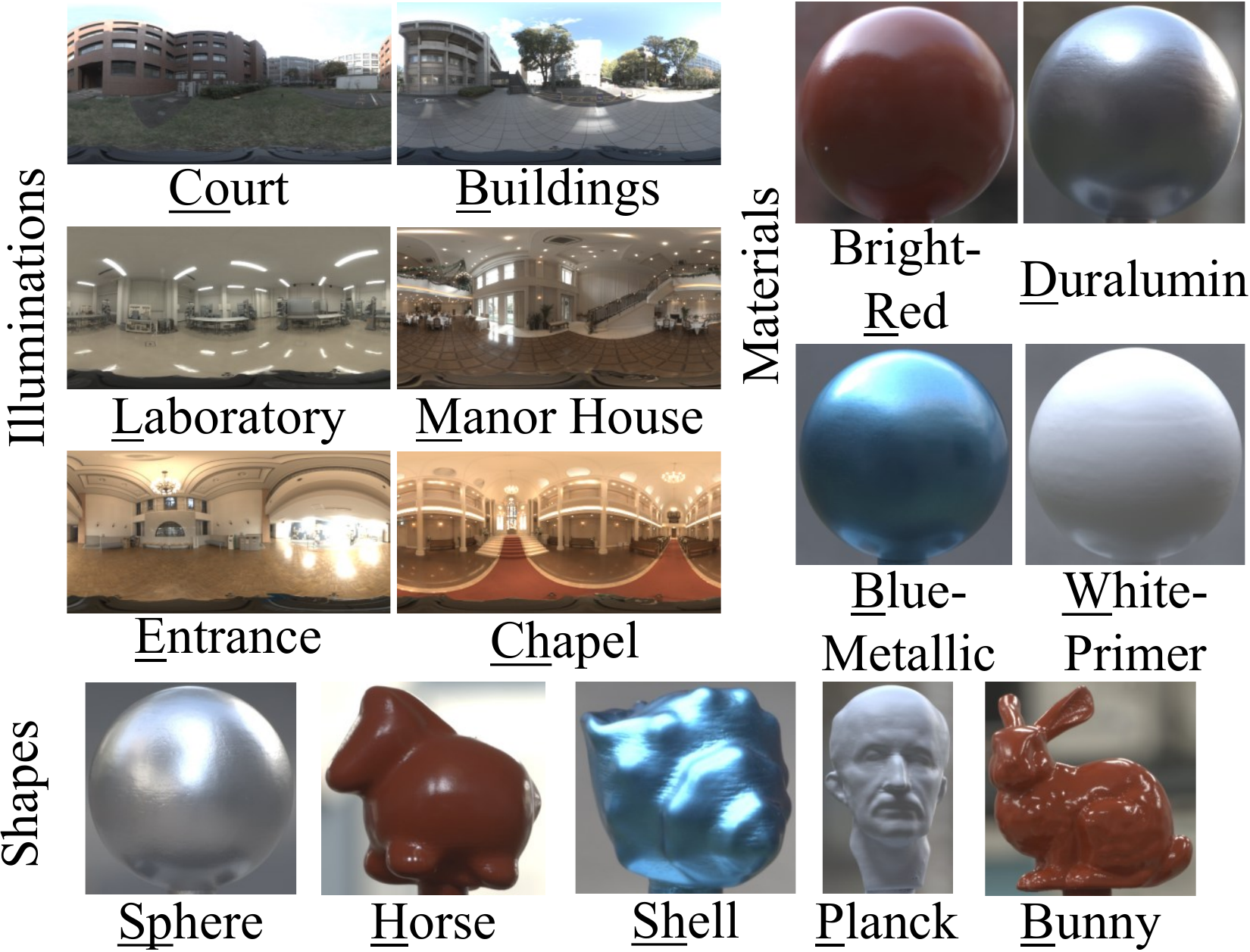}
  \caption{Our new multi-view real object image dataset, nLMVS-Real, consists of  radiometrically and geometrically accurately calibrated 2569 multi-view images of 120 combinations of 5 shapes, 4 materials, and 6 illumination environments.}
  \label{fig:nlmvs_real}
\end{figure}

We evaluate the effectiveness of nLMVS-Net through extensive experiments using both synthetic and real images of objects of different shapes and reflectances taken in a variety of illumination environments. For this, we introduce novel large-scale synthetic and real datasets, which we refer to as nLMVS-Synth and nLMVS-Real, respectively.

\begin{figure*}[t]
  \centering
  \includegraphics[keepaspectratio, width=\linewidth]{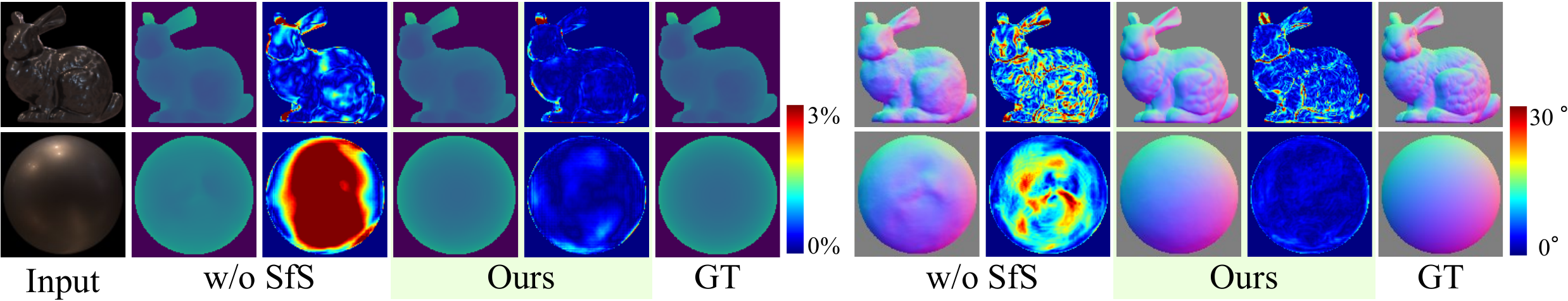}
   \caption{Ablation study on our shape-from-shading sub-network (``w/o SfS'') which constructs a cost volume from multi-view image features without leveraging radiometric cues as per-pixel probabilistic surface normal likelihoods. For each result, we show the estimation errors as a color map. The results clearly show that the shape-from-shading sub-network is essential to handle textureless, non-Lambertian objects.}
   \label{fig:result_ablation_sfs}
\end{figure*}

We first report that the Multiview Objects Under Natural Illumination Database \cite{oxholm2014mvshape} contains clear flaws in image capture including saturation, glare, and poor geometric calibration (please see the supplementary material for details). For this reason, numerical results on this dataset do not accurately reflect superiority of any method. This problem has been communicated with the authors of \cite{oxholm2014mvshape} and confirmed by them. In fact, one key contribution of our paper is the introduction of a new and a more extensive and accurate dataset that can replace this dataset. Radiometrically and geometrically accurate image capture for such dataset requires meticulous calibration and painstaking efforts in actual capture. Our new dataset (nLMVS-Real) would likely serve the community for a broad range of research on appearance modeling. 

\begin{table}[t]
  \centering
  \footnotesize
  \subfloat[][From 5 Views.]{
      \setlength{\tabcolsep}{3pt}
    \begin{tabular}{l|cr}
        & Depth & \multicolumn{1}{c}{Normal} \\ \hline
        RC-MVS \cite{chang2022rcmvsnet} & 5.76 \% & \multicolumn{1}{c}{-} \\
        MVSNeRF \cite{chen2021mvsnerf} & 5.59 \% & \multicolumn{1}{c}{-} \\
        CVP-MVS \cite{yang2020cvpmvsnet} & 4.16 \% & \multicolumn{1}{c}{-} \\
        IDR \cite{yariv2020idr} & 1.11\% & 11.0 deg  \\
        w/o SfS & 1.12 \% & 11.2 deg  \\
        \rowcolor[rgb]{0.93,1.0,0.87} Ours & 0.94 \% & 9.8 deg  \\
    \end{tabular}
    \label{tab:comparison}
  }
  ~~
  \subfloat[][From 10 Views.]{
  \vspace{4pt}
  \setlength{\tabcolsep}{3pt}
  \begin{tabular}{l|c}
    & Mesh \\ \hline
    NeRS \cite{zhang2021ners} & 0.63 \%  \\
    PhySG \cite{zhanf2021physg} & 0.61 \%  \\
    IDR \cite{yariv2020idr} & 0.25 \%  \\
    \rowcolor[rgb]{0.93,1.0,0.87} Ours + PSR \cite{kazhdan2006poisson} & 0.38 \% 
    \label{tab:result_mesh}
  \end{tabular}
  }

  \caption{\subref{tab:comparison} Mean errors of the estimated depths and surface normals on the nLMVS-Synth dataset. The results clearly show the effectiveness of our method. \subref{tab:result_mesh} Mean error of 3D mesh models recovered from 10 view images. Even though we recover the mesh models by simply applying Poisson surface reconstruction (PSR) \cite{kazhdan2006poisson}, our results are quantitatively comparable to the state-of-the-art methods.}
\end{table}

\vspace{-12pt}
\paragraph{nLMVS-Synth Dataset}
We rendered a large number of training and test images of synthetic shapes \cite{xu2018relighting, stanford3dscan, suggestivecontour, plyfiles}, measured BRDFs \cite{matusik2003brdf}, and captured illumination maps \cite{hdrihaven, gardner2017indoor, lightprobe, hiresolightprobe}. For training, we synthesized images of 2685 combinations of different shapes, materials, and illuminations. For each combination, we rendered images of randomly sampled 10 views. In total, the training set consists of 26,850 images. We also rendered a separate set of images for testing which consists of 4320 multi-view images of 216 different combinations of 6 shapes, 6 materials, and 6 illuminations. For each combination, we sampled 20 views on the horizontal line at equal intervals and added perturbations to them in the horizontal and vertical directions.

\vspace{-12pt}
\paragraph{nLMVS-Real Dataset}
\Cref{fig:nlmvs_real} shows example images from our new nLMVS-Real dataset. We captured approximately 20 HDR images at three different heights for each of all combinations of 5 shapes, 4 materials, and 6 illumination environments. We also captured illumination maps using RICOH THETA Z1. The objects are replicated using a 3D printer and painted with different materials. Ground-truth 3D mesh models are available for quantitative evaluations. 

\begin{figure*}[t]
  \centering
  \subfloat[][]{
    \includegraphics[keepaspectratio, height=136pt]{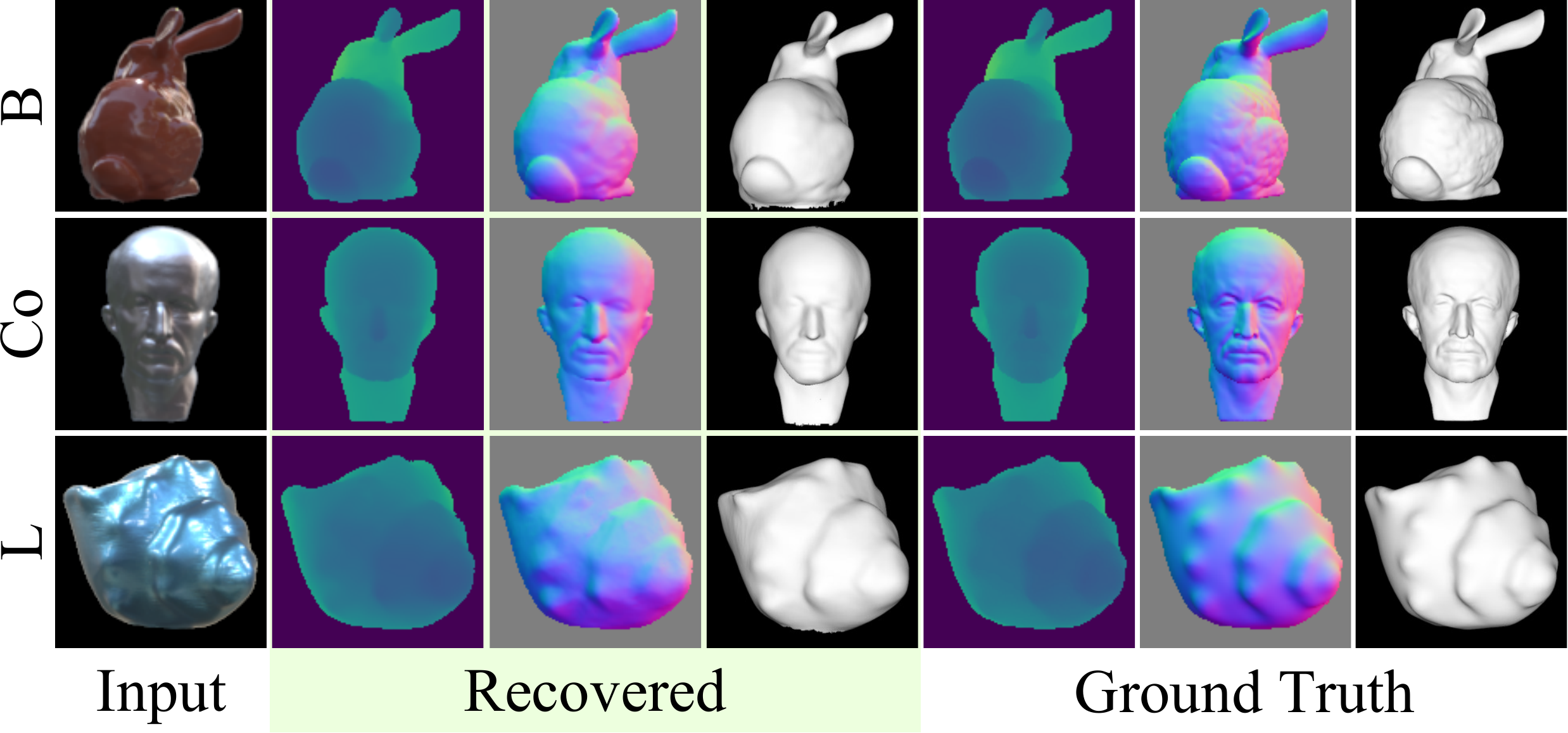}
    \label{fig:result_real_shape}
  }
  \quad
  \subfloat[][]{
    \includegraphics[keepaspectratio, height=136pt]{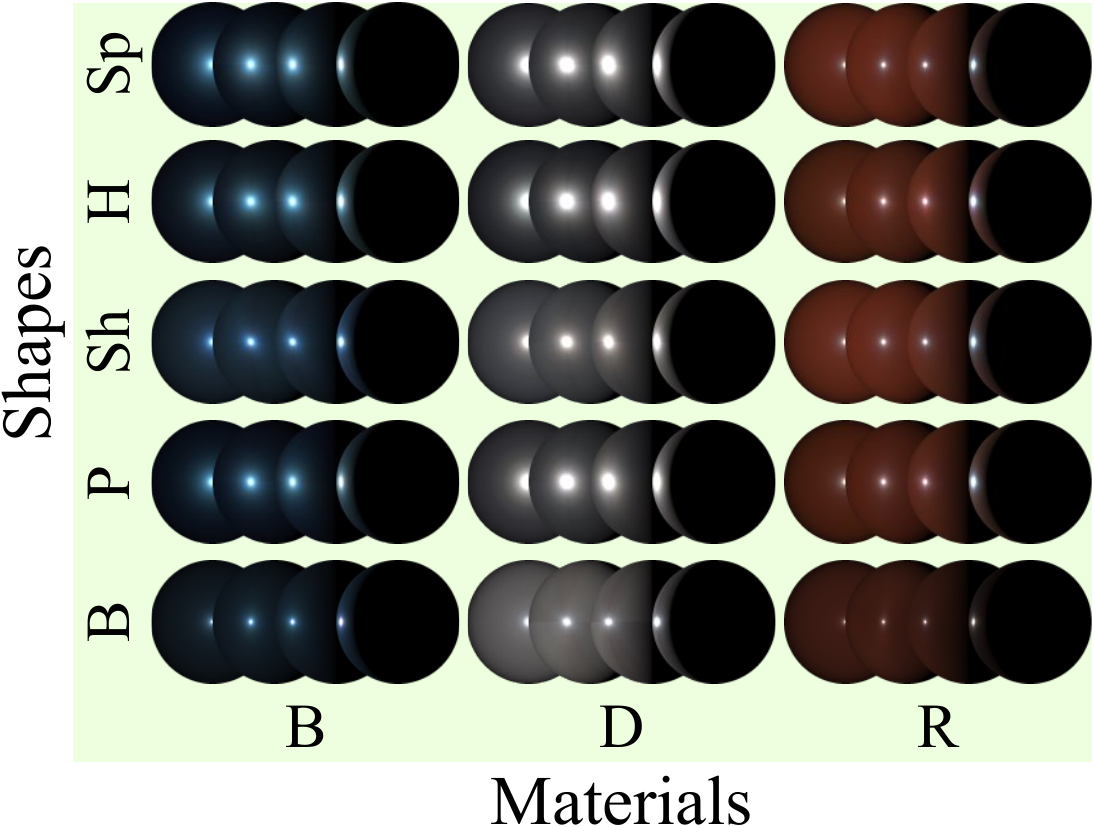}
    \label{fig:result_real_brdf}
  }
  \caption{\subref{fig:result_real_shape} Recovered depths, surface normals, and whole 3D shape (mesh models) from our nLMVS-Real dataset. Please see supp. material for more results. Letters on the left denote different illumination environments. Our method successfully recovers accurate geometry for a wide variety of real-world objects. \subref{fig:result_real_brdf} Estimated BRDFs under ``Court'' illumination. Please see supp. material for all results. The results are consistent across estimation from images of different shapes, which demonstrates the accuracy of our method. }
\end{figure*}

\vspace{-12pt}
\paragraph{Baseline Methods}
We compare our method with CVP-MVSNet \cite{yang2020cvpmvsnet}, MVSNeRF \cite{chen2021mvsnerf}, and RC-MVSNet \cite{chang2022rcmvsnet} which also handle sparse (\ie, 3 or 5 view) inputs. We could not directly compare our method with Kusupati \etal \cite{kusupati2020nas} as their network is trained to recover depths and normals for the entire scene rather than a single object. Instead, we compare our method with ours ``w/o SfS'' (without shape-from-shading) that constructs a cost volume only from multi-view image features similar to Kusupati \etal \cite{kusupati2020nas}. We also compare our method with IDR \cite{yariv2020idr}, a neural image synthesis method that recovers surface geometry from relatively sparse (\ie, typically 50 and 11 at minimum) view inputs. 

\vspace{-12pt}
\paragraph{Evaluation Metrics}
We measure the accuracy of the recovered depth and surface normals with mean absolute errors. For depth error, the scale of the object is normalized such that the diagonal length of its bounding box is 1.

\subsection{Results on Synthetic Data}

\paragraph{Accuracy of the Shape-from-Shading Network}
We first evaluate the accuracy of the proposed shape-from-shading network. Ground-truth BRDF was used to compute the input surface normal likelihood; \ie, we assume that the object's reflectance is known. The error between the ground-truth surface normals and those estimated to have the highest probability was lower than 10 degrees for 83\% of all pixels. This is comparable to the accuracy of existing shape-from-shading methods such as Johnson and Adelson \cite{johnson11sfs} and the single-view method of Oxholm and Nishino \cite{oxholm2015shape}.

\vspace{-12pt}
\paragraph{Joint Shape and Reflectance Estimation Results}
\Cref{fig:opening,fig:result_ablation_sfs}, and \cref{tab:comparison} show qualitative and quantitative results. While existing methods and our method without the shape-from-shading cues (\ie, ``w/o SfS'') fail on non-Lambertian and textureless objects, our method successfully recovers both depths and surface normal for these challenging objects. For 93 \% of all input images, mean depth and surface normal errors were lower than 2 \% and 19 degrees, respectively. These clearly show the effectiveness of our method. Please see the supplementary material for more results and ablation studies.

\vspace{-12pt}
\paragraph{Accuracy of the Whole 3D Shape Recovery}
We can also recover 3D mesh models from our depth and surface normal estimates of 10 views (\cref{sec:whole_shape_recovery}) and compare the results with those of neural image synthesis methods \cite{zhang2021ners, zhanf2021physg, yariv2020idr}. For this experiment, we used 10 views uniformly sampled from the original 20 views of the nLMVS-Synth dataset. We evaluate the reconstruction accuracy with root-mean-square (RMS) of the distance from a point on the reconstructed mesh to the nearest point on the ground truth mesh. \Cref{tab:result_mesh} shows quantitative results.  Even though we recover the mesh models by simply applying Poisson surface reconstruction \cite{kazhdan2006poisson}, our results are quantitatively comparable to the state-of-the-art methods. Please see the supplementary material for qualitative results.

\subsection{Results on Real Data}

\Cref{fig:result_real_shape} shows qualitative results of the recovered geometry on our nLMVS-Real Dataset. The results are of high quality even for complex shapes. Mean depth and surface normal errors were 2.01 \% and 13.6 degrees, respectively. For 70 \% of all input images, depth error was lower than 2 \% and surface normal error was lower than 17 degrees. \Cref{fig:result_real_brdf} shows several of the estimated BRDFs. The estimates are consistent across different shapes. Note that ground truth BRDF of the real materials cannot be easily acquired. 

As we make several assumptions about the objects and the capturing setup (\eg, homogeneous material and distant illumination), the estimation accuracy would decrease for large deviations from these assumptions. Nevertheless, as the experimental results show, our method successfully recovers geometry and reflectance from real-world images that do not strictly satisfy them, which demonstrates the robustness of our method.

\section{Conclusion}
In this paper, we introduced nLMVS-Net, a neural multi-view stereo network that can recover both depth and surface normal at each pixel in the reference view for objects with complex reflectance taken under known but natural illumination. The method integrates radiometric cues in the form of view-independent surface normals recovered with a dedicated network into depth and surface normal cost volume filtering. By canonically modeling uncertainties of the surface normals, they provide rich cues for accurate geometry recovery. Experimental results clearly demonstrate the effectiveness of nLMVS-Net including its accuracy in recovering the complex reflectance of real-world objects. We believe nLMVS-Net can serve as a useful practical means for passive geometry recovery in the wild.

\vspace{-6pt}
\paragraph*{Acknowledgement}
This work was in part supported by
JSPS 
20H05951, 
21H04893, 
JST
JPMJCR20G7, 
JPMJSP2110, 
and RIKEN GRP.
We also thank Shinsaku Hiura for his help in 3D printing.

{\small
\bibliographystyle{ieee_fullname}
\bibliography{wacv23final_nlmvs}
}

\end{document}